\RequirePackage{amsmath}
\documentclass[runningheads]{llncs}
\usepackage[T1]{fontenc}
\usepackage{graphicx}
\usepackage{hyperref}

\usepackage{color}
\usepackage{xcolor}
\usepackage{tikz}
\definecolor{pltgreen}{HTML}{4daf4a}

\usepackage{cleveref}
\usepackage{booktabs}
\usepackage{amssymb}

\begin{document}
\title{Dancing to the State of the Art?}
\subtitle{How Candidate Lists Influence LKH for Solving the Traveling Salesperson Problem}
%
%
\author{
Jonathan~Heins\inst{1}\thanks{Equal contributions.}\orcidID{0000-0002-3571-667X} \and
Lennart~Schäpermeier\inst{1,2}$^\star$\orcidID{0000-0003-3929-7465} \and
Pascal~Kerschke\inst{1,2}\orcidID{0000-0003-2862-1418} \and
Darrell~Whitley\inst{3}\orcidID{0000-0002-2752-6534}
}
\authorrunning{J. Heins et al.}
%
\institute{
Big Data Analytics in Transportation, TU Dresden, Germany
\email{\{jonathan.heins,lennart.schaepermeier,pascal.kerschke\}@tu-dresden.de}
\and
ScaDS.AI Dresden/Leipzig, Dresden, Germany
\and
Department of Computer Science, Colorado State University, Fort Collins, USA
\email{whitley@cs.colostate.edu}
}
\maketitle              
\begin{abstract}
Solving the Traveling Salesperson Problem (TSP) remains a persistent challenge, despite its fundamental role in numerous generalized applications in modern contexts. Heuristic solvers address the demand for finding high-quality solutions efficiently. Among these solvers, the Lin-Kernighan-Helsgaun (LKH) heuristic stands out, as it complements the performance of genetic algorithms across a diverse range of problem instances. However, frequent timeouts on challenging instances hinder the practical applicability of the solver. 

Within this work, we investigate a previously overlooked factor contributing to many timeouts: The use of a fixed candidate set based on a tree structure. Our investigations reveal that candidate sets based on Hamiltonian circuits contain more optimal edges. We thus propose to integrate this promising initialization strategy, in the form of POPMUSIC, within an efficient restart version of LKH. As confirmed by our experimental studies, this refined TSP heuristic is much more efficient -- causing fewer timeouts and improving the performance (in terms of penalized average runtime) by an order of magnitude -- and thereby challenges the state of the art in TSP solving.

\end{abstract}

\section{Introduction}

The \textit{Traveling Salesperson Problem (TSP)}, also referred to as \textit{Traveling Salesman Problem} or \textit{Traveling Sales-rep Problem}, is one of the most well-known combinatorial optimization problems and has thus been studied and researched in various domains for decades. While it often serves as a motivating example in various computer science, transportation science, and logistics courses and textbooks, it is also of practical use in less obvious disciplines such as biology (e.g., in the context of DNA sequencing) or astronomy \cite{applegate2011traveling}. Furthermore, it is fundamental to various related, albeit potentially more realistic, variants of routing-based optimization problems such as the \textit{Vehicle Routing Problem (VRP)} with its numerous extensions or modifications of its own, or capacity-constrained problems such as the \textit{Traveling Thief Problem (TTP)}.

Essentially, the TSP is a graph-based combinatorial optimization problem, which belongs to the class of $\mathcal{NP}$-hard problems. Given a complete and undirected graph $G = (V, E)$ with a set of \textit{vertices} $V = \{1, \ldots, n\}$ (also called \textit{cities} or \textit{nodes}), and a set of \textit{edges} $E \subseteq V \times V$ connecting each pair of vertices $\{i, j\} \in E$, we seek the optimal round trip along a subset of all available edges $E$ such that each node is visited exactly once. While the previous requirements already suffice to define the TSP, one usually focuses on a particular variant of it, the \textit{Euclidean TSP}, which is also focused in this work. As an extension to the previous definition of the TSP, the Euclidean TSP fulfills two additional properties: First, the cost $c: E \rightarrow \mathbb{R}$ for traveling along an edge is independent of the direction of travel. 
Secondly, the previously mentioned costs per edge correspond to the Euclidean distance between the two nodes that are connected by the respective edge.

Approaches for solving TSPs are typically divided into two types: exact and inexact methods. The former have the benefit of guaranteeing the optimality of the found solution, once the algorithm terminated successfully. In this class of algorithms, Concorde~\cite{applegate2011traveling} is regarded as the unanimous state of the art. In contrast, inexact heuristics -- such as EAX \cite{nagata1997edge,nagata2013powerful}, LKH \cite{helsgaun2000effective,helsgaun2009general}, Mixing GA \cite{varadarajan2019}, or MAOS \cite{xie2008multiagent} -- have the advantage that they are generally much faster at finding high-quality solutions. However, they lack any optimality guarantees for their found solutions.
There is no single state-of-the-art heuristic for inexact TSP solving, as indicated by the portfolio listed above. Yet, numerous benchmark and algorithm selection studies \cite{kotthoff2015improving,kerschke2018parameterization,heins2023} have shown a general trend towards the superiority of EAX and LKH. This trend became even more evident since the refinement of both heuristics with external restart mechanisms \cite{duboislacoste2015}.

Interestingly, for both heuristics, EAX and LKH, the restart mechanism clearly reduced the number of timeouts (i.e., the algorithm runs that did not find the optimal tour within the given time budget), with the effect being considerably more apparent for EAX. A possible reason for the weaker impact of the restart mechanism on LKH could be an insufficient variety among the solutions that are initially generated after each restart. This hypothesis is also supported by two recent studies, which have shown that (a) LKH is very sensitive to its hyperparameter configurations \cite{seiler2023}, and (b) as long as the number of 2-opt runs (which are an essential component of LKH) is sufficiently large, one can ensure that all edges of the optimal tour have been found \cite{varadarajan2019}.

Despite the numerous benchmark studies, too little is known about the strengths and weaknesses of EAX and LKH -- in particular, which structures of TSP instances are easy for one and hard for the other. To better understand the effects of certain node alignments on the performance of EAX and LKH, \cite{bossek2019} created several sets of artificial TSP instances that are easy for EAX but hard for LKH and vice versa. They proposed several mutation operators that transform a given TSP instance according to topological patterns such as grids, lines, explosions, implosions, or clusters. They then integrated these operators into an evolutionary algorithm to evolve TSP instances that are easy to solve for one heuristic but challenging for the other. It is striking that it was relatively easy to generate EAX-friendly instances that LKH could not solve. In contrast, although EAX solved LKH-friendly instances more slowly than LKH, it could still find solutions quickly. Although several feature-based studies have investigated these effects in recent years \cite{heins2021,heins2023,kerschke2018leveraging}, one still needs to find an answer for why and when LKH has difficulties with some of these instances.

A likely reason for the occasional timeouts and sub-par performance of LKH could be the approach that produces its candidate sets. If the 
edges from the candidate sets do not already contain all edges of the optimal tour, LKH appears to have difficulties combining the candidates into the optimal tour. However, the recent integration of the metaheuristic POPMUSIC \cite{taillard1993parallel,ribeiro2002popmusic} into LKH \cite{taillard2019popmusic} could resolve this issue. It shows the potential to generate more diverse initial candidate sets than the classical initialization approach of LKH, thereby increasing the probability that all relevant edges are included among the candidates. 

So far, the integration of POPMUSIC in LKH has mainly been demonstrated for large TSP problems with more than 10\,000 nodes per instance \cite{taillard2019popmusic}. However, as we will show in this work, POPMUSIC can also leverage the performance of LKH on small- to medium-sized TSP instances with up to 2\,000 nodes. Furthermore, the effectiveness of POPMUSIC has so far only been examined in combination with the classical restart-free implementation of LKH. For this work, we have implemented POPMUSIC in a more effective restart version of LKH. As our results will show, combining the two components leverages the power of both, ultimately resulting in a powerful TSP heuristic that can challenge the state-of-the-art status of the restart-version of EAX in inexact TSP solving.

The remainder of this work is structured as follows. Section~\ref{sec:background} describes the key concepts of LKH with a dedicated focus on the strategies for producing its candidate sets. The weaknesses of LKH's default strategy, the $\alpha$-candidate set, are discussed in Section~\ref{sec:alpha}. The experimental setup of our study is described in Section~\ref{sec:setup} and its results are analyzed in Section~\ref{sec:results}. At last, Section~\ref{sec:conclusion} summarizes our work and presents ideas for future research that are enabled by our work.

\section{Background}
\label{sec:background}

The Lin-Kernighan Heuristic \cite{lin1973effective} dates back to 1973 and is still a highly competitive algorithm in the form of LKH.
The core concept introduced by Lin and Kernighan  behind the heuristic has not changed and revolves around the construction of sequential $k$-opt moves, where $k$ is unrestricted. 
If $k$ equals the number of cities in a TSP instance, a $k$-opt move could potentially yield the optimal solution directly from any feasible solution. 
However, a \textit{sequential} $k$-opt move is restricted by allowing only moves that can be constructed through consecutive basic moves, e.g., 2-opt moves. 
After each such basic move, the resulting edges must form a Hamiltonian circuit. 
Despite this restriction, it is computationally infeasible to evaluate every possible edge for each basic move.
The primary strategy that LKH employs to reduce this computational complexity involves pruning the search space.
The following sections will introduce the different pruning strategies.

\subsection{$\alpha$-Candidate Set}

In its original form, the Lin-Kernighan (LK) algorithm \cite{lin1973effective} constrained the search by considering only edges connecting each vertex to one of its five closest neighbors. 
However, Helsgaun \cite{helsgaun2000effective} identified instances where the optimal edges did not emerge from this restricted set of candidates generated by the five-nearest-neighbor heuristic.
Consequently, LKH incorporates a more sophisticated approach, leveraging minimum 1-trees to derive a more accurate approximation of edge proximity to the optimal tour.
Additionally, it employs subgradient optimization techniques to further refine these estimations, enhancing the algorithm's effectiveness in finding high-quality solutions.

\begin{figure}[tb]
\begin{minipage}[t]{0.41\textwidth}
  \centering
  \includegraphics[height = 4.25cm, keepaspectratio]{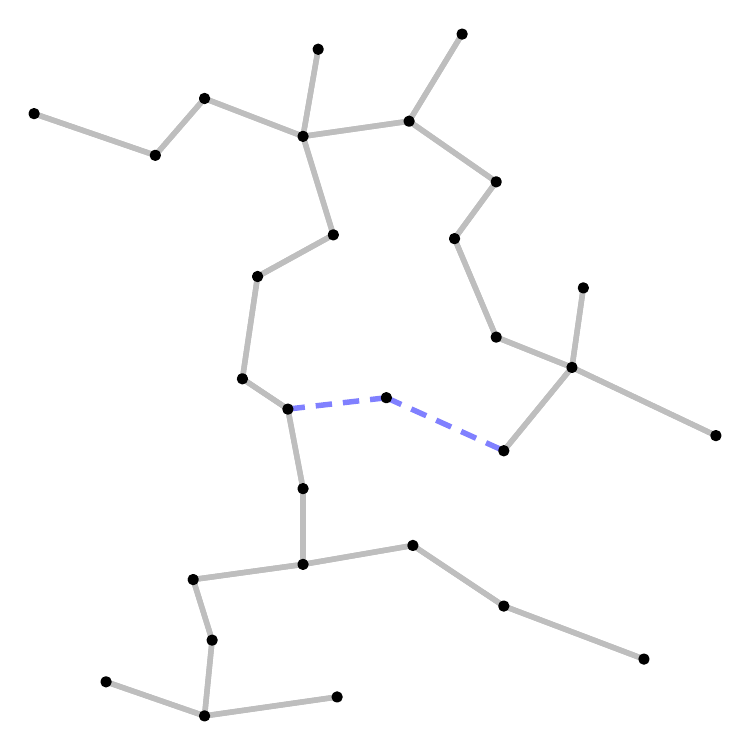}
  \vspace*{-3pt}
  \caption{A base minimum 1-tree is composed of an MST (gray edges), along with two additional (blue dashed) edges connecting the excluded node with its first and second nearest neighbors.}
  \label{fig:m1tree-base}
\end{minipage}
\hfill
\begin{minipage}[t]{0.52\textwidth}
  \centering
  \includegraphics[height = 4.25cm, keepaspectratio]{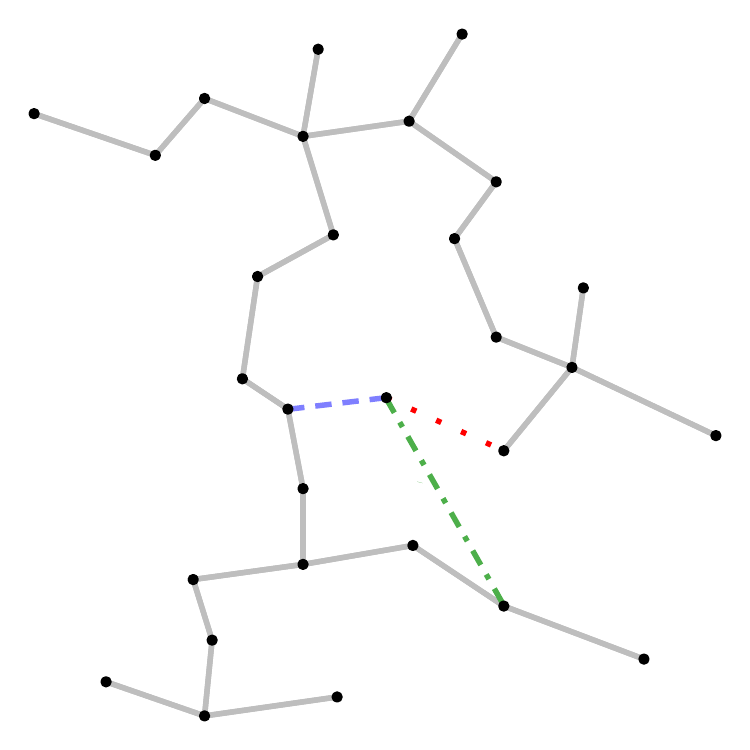}
  \vspace*{-3pt}
  \caption{The minimum 1-tree required to include an additional (green dashed-dotted) edge does not have to include the second nearest neighbor edge (red dotted line). Thus, the $\alpha$-nearness value of the new edge is the difference between itself and the red edge.}
  \label{fig:m1tree-reqedge}
\end{minipage}
\vspace*{-3pt}
\end{figure}

A minimum 1-tree is related to, but distinct from, a spanning tree; it is formed by a minimum spanning tree encompassing all nodes except one, which is arbitrarily chosen. 
This remaining node is linked to the tree via two edges, connecting it with its closest two neighboring nodes, thus introducing a cycle (see \Cref{fig:m1tree-base}). 
If every vertex in this structure has a degree of two, the minimum 1-tree manifests as a Hamiltonian circuit, effectively resolving the TSP. 
Helsgaun's investigation revealed that a typical minimum 1-tree encompasses approximately $70 \%$ to $80 \%$ of the edges present in the optimal tour. 
Consequently, he introduced an $\alpha$-nearness metric, which computes the cost associated with incorporating each edge into the minimum 1-tree. 
This metric quantifies the delta between a base minimum 1-tree (as depicted in \Cref{fig:m1tree-base}) and an adjusted minimum 1-tree required to accommodate the edge under consideration (see \Cref{fig:m1tree-reqedge}).

Adjusting the distance matrix by adding a constant to the distances to and from one node -- i.e., updating the corresponding row and column in the distance matrix -- can modify the minimum 1-tree while preserving the optimal tour. 
In the optimal tour, every node has to be connected with two edges, thus there cannot be any benefit in changing node permutation if all distances from or to one node are changed. 
In the case of a minimum 1-tree a node can be connected with up to six edges, see \cite{heins2021,heins2023} a more detailed explanation with minimal spanning trees. 
Increasing the distance from one node to its nearest neighbors will thus lead to fewer connections to this node in a minimum spanning tree. 
With this, a refinement through subgradient optimization can be applied such that the minimum 1-tree more closely resembles a tour, via weights per node added to the distances of all edges including the node \cite{subgradient1,subgradient2,helsgaun2000effective}. 
Essentially, the subgradient algorithm iteratively increases the weights, i.e., the distance in the distance matrix, of nodes with a degree of 3 or more and decreases the weights of leaf nodes within the minimum 1-tree. 
When constructing a new minimum 1-tree with the altered distance matrix, the added or subtracted constants in tendency lead to previously higher degree nodes being connected with fewer other nodes and previous nodes of degree one being connected with more nodes. 
This optimization process is integrated into LKH and terminates either upon finding the optimal tour or after not finding an improving minimum 1-tree for too many iterations. 
The resulting minimum 1-tree is reported to yield ranks in terms of smallest $\alpha$ values for the optimal edges that are on average $1.7$, compared to $2.1$ based on the non-transformed minimum 1-tree \cite{helsgaun2000effective}. 
Note that a rank of $1.5$ would be optimal as every node is part of two optimal edges which should in the optimal case have the two smallest $\alpha$ values.

\subsection{2-opt Candidate Set}

The use of minimum spanning trees is indeed likely to find edges with low cost that connect one vertex to another.
But this strategy does not take into account that the edges must be part of a Hamiltonian circuit.
By constructing locally optimal Hamiltonian circuits to sample edges, we improve the probability that the sampled edges can contribute to 
the 
globally optimal solution. 
Also, minimum spanning trees do not guarantee that there are at least two edges that are incident on every vertex. For example,
in an asymmetric TSP,  one directed edge must ``enter'' the city,
and one directed edge must ``leave'' the city.   This same idea
generalizes to symmetric TSP instances.
We thus argue that a more obvious and likely more productive way to initialize and populate the candidate set is to use edges that appear in some locally optimal Hamiltonian circuit.

One way to find useful edges is to sample them from local optima.
A standard way to both define and discover local optima is to use the 2-opt operators.
The 2-opt operator, when applied until no further improvement is possible, is guaranteed to generate locally optimal Hamiltonian circuits without any crossing edges \cite{croes1958method}.
This guarantee only holds, however, if the TSP instance uses a cost matrix that obeys the triangle inequality.
The triangle inequality holds for 
all Euclidean TSP benchmark problems, 
which are the focus of this work.

The successful and effective EAX genetic algorithm \cite{nagata2013powerful} uses an efficient implementation of 2-opt. It has been shown to generate a set of locally optimal solutions that sample edges found in the global optimum at a very high rate. 
Usually, the edges found in these local optima are also found in the global optimum at a rate of approximately 70\% \cite{varadarajan2019}.  

\Cref{tab:pop_size} presents a sample data that was previously presented by Varadarajan et al.~\cite{varadarajan2019}.
The problems are a mixture of clustered TSP instances, printed circuit board (PCB) problems, and ``city problems'' based on the coordinates of cities located in various countries.
The key information found in this table is the population size necessary to ensure (empirically) that all of the edges found in the global optimum were also present in the population.
In almost all cases they examined, a population size of 256 was sufficient.
The exceptions are shown in \Cref{tab:pop_size}: C3k.1 and vm1084 required a population of 512, instance fl1577 required a population of 4096, and finally d2103 required a population of 16,384.
We should note that some of these problems (att532 and u1817) are known to have multiple global optima.
The listed results only checked for edges in one of the global optima (arbitrarily chosen).

\begin{table}[tb]
\centering
  \caption{The following sample of problem instances was previously used to show the
  effectiveness of 2-opt at initializing a population for a genetic algorithm.
  The table records the smallest population size which contains all of the edges found 
  in the global optimum after performing 2-opt. See \cite{varadarajan2019} for the complete table.}
  \label{tab:pop_size}
  \begin{tabular}{cc|cc|cc}
    \toprule
    Cluster & Pop & PCB & Pop & City & Pop \\
    ~~Problem~~ & ~~~~Size~~~~ & ~~Problem~~ & ~~~~Size~~~~ & ~~Problem~~ & ~~~~Size~~~~ \\
    \midrule   
   C1k.0 & 256 &  d657 & 256 & att532 & 64 \\
   C3k.0 & 256 & pcb1173 & 256 & pr1002 & 128 \\
   C3k.1 & 512 & d1291 & 128 & vm1084 & 512 \\
 dsj1000 & 128 	& fl1577 & 4096 &  rl1323 & 256 \\
   & &  d1655 & 256 & nrw1379 & 128 \\
   & & u1817 & 128 &  vm1748 & 128 \\
  & &  d2103 & 16384 & rl1889 & 128 \\
   & & u2319 & 64 & pr2392 & 128 \\
   & & pcb3038 & 128 & fnl4461 & 256 \\
  \bottomrule
\end{tabular}
\end{table}

The significance of these results is that they suggest that 2-opt might be used to initialize a population of locally optimal solutions, and the edges in those solutions could be used to construct a candidate set where all of the edges are known to occur in a Hamiltonian circuit that is also locally optimal.  

\subsection{POPMUSIC Candidate Set}

Another similar approach that has been directly implemented in LKH (since version 2.0.9) is \textit{Partial OPtimization Metaheuristic Under Special Intensification Conditions (POPMUSIC)} for TSP \cite{taillard2019popmusic}. 
In a nutshell, POPMUSIC optimizes an initial tour by optimizing sub-paths of consecutive cities on the current tour at each iteration.
The initial tour is found by starting with an LK-optimized tour on a subsample of the cities. 
Then, it adds the out-of-tour nodes to the tour, which is/are located closest to the tour. 
For each of the cities added in this second step, it then optimizes the sub-path around it using 2-opt. 
Finally, sub-paths of consecutive cities are optimized until no further improvement can be found. 
For further details, we refer to the original paper \cite{taillard2019popmusic}.

This procedure was originally developed to efficiently generate edge candidate sets for large instances, where the $\alpha$-set generation requires too much time.
The authors of \cite{taillard2019popmusic} note that, while the costs of tours generated by POPMUSIC are not particularly good, multiple runs tend to include (almost) all optimal edges, making it suited for candidate set generation.

In our preliminary experiments, the quality of edges produced by POPMUSIC and 2-opt is similar between both approaches.
Due to the better accessibility and scalability in a restart variant, we generally recommend using the built-in POPMUSIC technique rather than using custom 2-opt tours as the basis 
for Hamiltonian circuit-based candidates.

\section{$\alpha$-Candidate Set Pitfalls}
\label{sec:alpha}

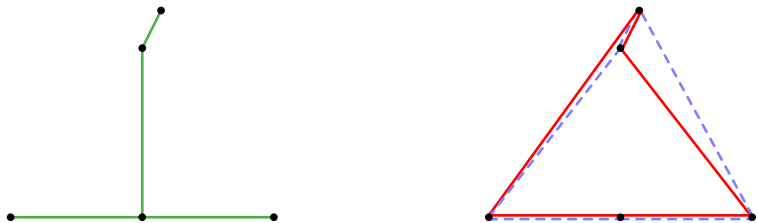
\begin{figure}[t]

\begin{minipage}[t]{0.47\textwidth}

    \centering
    \begin{tikzpicture} [line cap=round,x=1.0cm,y=1.0cm, scale=2.5]
    \clip(-0.75,-0.3) rectangle (0.75,1.15);
    \draw [line width=1.pt, color=pltgreen] (0., 0.0)-- (0.7, 0.0);
    \draw [line width=1.pt, color=pltgreen] (0., 0.0)-- (-0.7, 0.0);
    \draw [line width=1.pt, color=pltgreen] (0., 0.0)-- (0.0, 0.9);
    \draw [line width=1.pt, color=pltgreen] (0.0,0.9)-- (0.1,1.1);
    \begin{scriptsize}
    \begin{scope}[every node/.style={draw, circle}]
    \draw [fill=black]  (0.,0.) circle          (0.5pt);
    \draw [fill=black]  (0.,0.9) circle         (0.5pt);
    \draw [fill=black] (0.1,1.1) circle         (0.5pt);
    \draw [fill=black] (0.7,0.) circle          (0.5pt);
    \draw [fill=black] (-0.7,0.) circle         (0.5pt);
    \end{scope}
    \end{scriptsize}

    \end{tikzpicture}
        
\end{minipage}
\hfill
\begin{minipage}[t]{0.47\textwidth}

    \centering
    \begin{tikzpicture} [line cap=round,x=1.0cm,y=1.0cm, scale=2.5]
    \clip(-0.75,-0.3) rectangle (0.75,1.15);
    \draw [line width=1.pt, color=red] (0., 0.01)-- (0.7, 0.01);
    \draw [line width=1.pt, color=blue!50, dashed] (0., -0.01)-- (0.7, -0.01);
    \draw [line width=1.pt, color=red] (0.7, 0.)-- (0.,0.9);
    \draw [line width=1.pt, color=blue!50, dashed] (0.7, 0.01)-- (0.1,1.11);
    \draw [line width=1.pt, color=red] (0.00894,0.8955)-- (0.10894,1.0955);
    \draw [line width=1.pt, color=blue!50, dashed] (-0.00894,0.9045)-- (0.09106,1.1045);
    \draw [line width=1.pt, color=red] (0.1,1.11)-- (-0.7,0.01);
    \draw [line width=1.pt, color=blue!50, dashed] (0.,0.9)-- (-0.7,0.);
    \draw [line width=1.pt, color=red] (-0.7,0.01)-- (0.,0.01);
    \draw [line width=1.pt, color=blue!50, dashed] (-0.7,-0.01)-- (0.,-0.01);
    \begin{scriptsize}
    \begin{scope}[every node/.style={draw, circle}]
    \draw [fill=black]  (0.,0.) circle                                 (0.5pt);
    \draw [fill=black]  (0.,0.9) circle                                   (0.5pt);
    \draw [fill=black] (0.1,1.1) circle                                 (0.5pt);
    \draw [fill=black] (0.7,0.) circle                               (0.5pt);
    \draw [fill=black] (-0.7,0.) circle                             (0.5pt);
    \end{scope}
    \end{scriptsize}

    \end{tikzpicture}
        
\end{minipage}
\vspace*{-18pt}
\caption{A minimalistic example of a TSP instance (with five nodes) that is not solvable by LKH. Left: The corresponding $\alpha$-candidate set. Right: The initial tour (solid red lines) with two missing edges from the optimal tour (blue dashed lines). }
\label{fig:slkhfail}
\vspace*{-6pt}
\end{figure}

Independent of the pruning strategy, LKH's ability to generate good solutions is limited when too many edges are removed from the search space. 
In fact, there are only three scenarios in which an edge, that is not part of the candidate set, can be introduced into the tour: (i) during the construction of the initial tour, (ii) to close a sequential $k$-opt move, and (iii) when \textit{kicking} a local optimal tour. 
Kicking a tour refers to altering parts of a tour to escape local optima after no improving move can be found anymore. 
Helsgaun's strategy is to create a new random tour and disallow removing
edges which are part of the current best solution \cite{helsgaun2009general}.
An alternative strategy is a double bridge move, which combines two 2-opt moves that, if applied alone, would result in two separate circles.

The limitation to only three scenarios, which enable introducing edges that are not yet part of the candidate set, can even be seen for very small TSP instances as depicted in \Cref{fig:slkhfail}. 
Although this instance consists of only five nodes, LKH cannot solve it, if the following parameters are set: disabling the subgradient procedure (\texttt{SUBGRADIENT = NO}), allowing only one candidate per city (\texttt{MAX\_CANDIDATES = 1}), changing the kick-strategy to a double bridge move (\texttt{KICK\_TYPE = 4}), and using the initial tour shown in \Cref{fig:slkhfail}.  
The tree-based candidate set as visible in \Cref{fig:slkhfail} lacks two edges crucial to the optimal tour, preventing their simultaneous introduction with scenario (ii). 
Additionally, since these edges are absent in the initial tour, and since a double bridge move changes four edges, they cannot be introduced during scenarios (i) and (iii). 
In this case of a very small search space with only five cities, LKH's kicking-strategy could by chance result in the optimal solution.
This is not the case with the double bridge move employed here.
When LKH attempts to replace a suboptimal edge, the only viable replacement is the middle vertical edge, which cannot be part of the optimal tour since it would require breaking either the left or right horizontal edge. 
Since the candidate set is based on a tree structure and not a Hamiltonian circuit, crucial edges can only be included if the number of candidates per point is increased. 
However, this increase can be counteracted by clustering many points near the existing ones.

\begin{figure}[tb]
    \centering
    \includegraphics[width=0.495\linewidth]{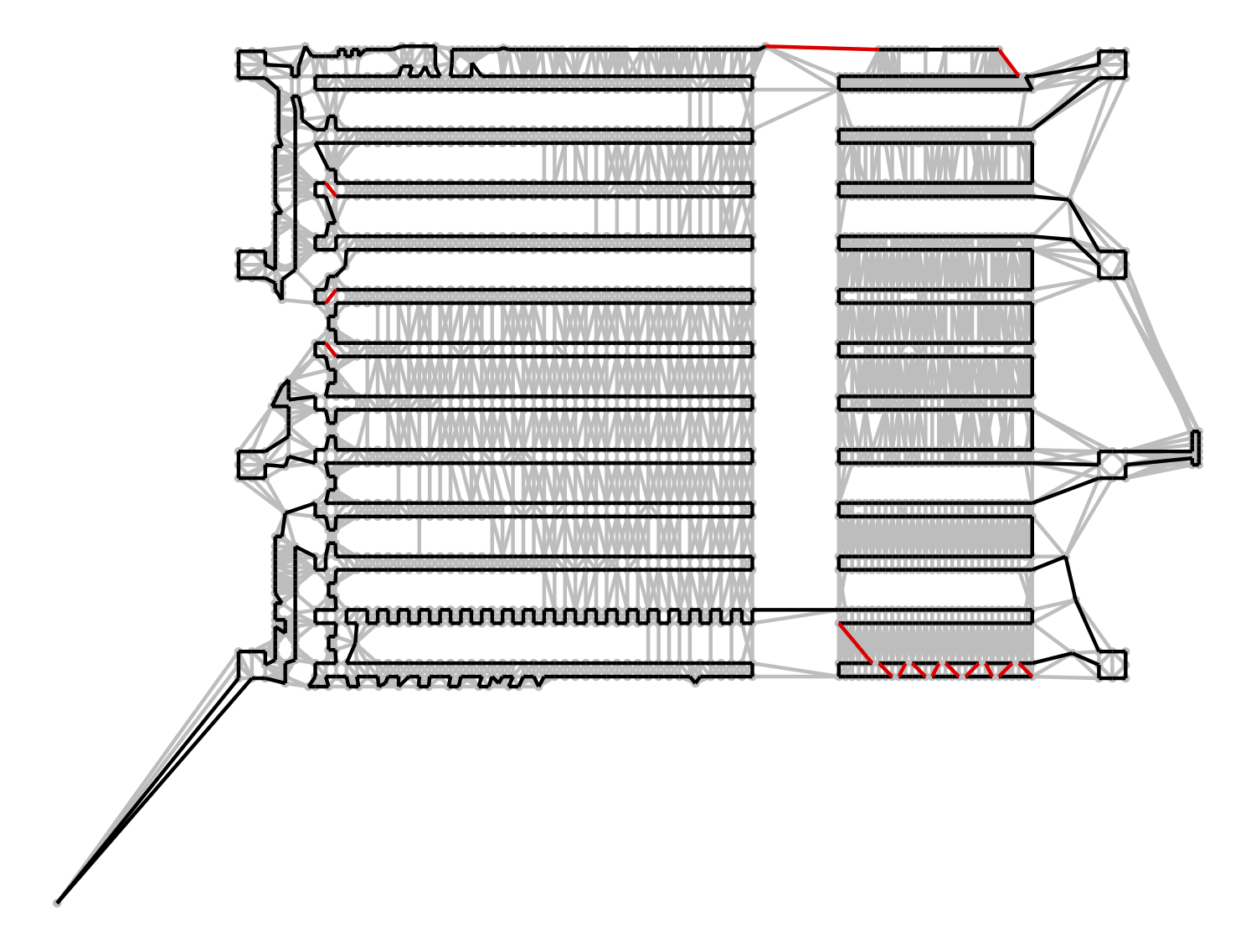}
    \includegraphics[width=0.495\linewidth]{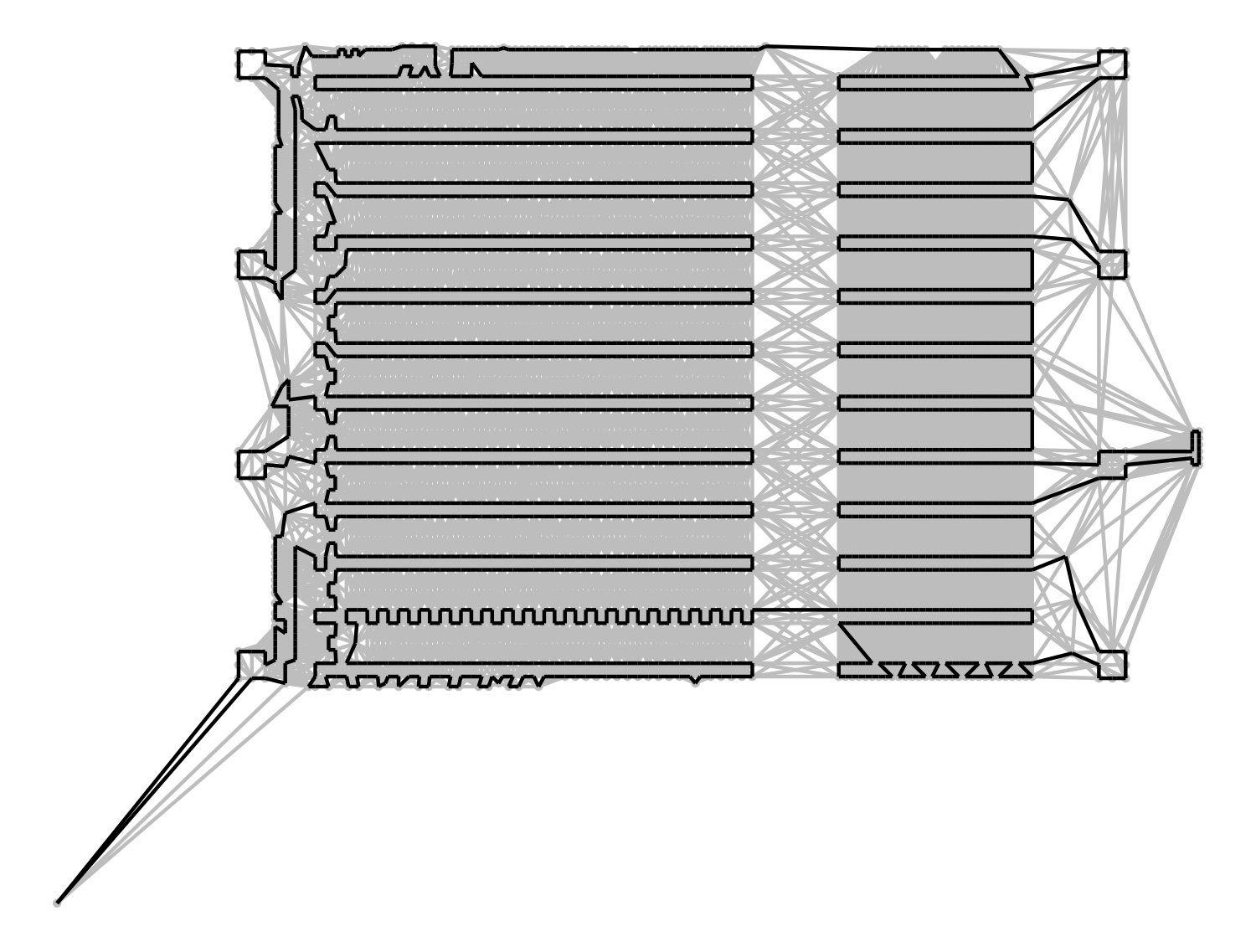}
    \vspace*{-12pt}
    \caption{$\alpha$-candidate set (left) and combined edges of $1,\!000$ 2-opt runs (right) with the optimal tour for \texttt{tsplib} instance \texttt{d2103}. Gray edges are not in the optimal tour, while black edges are part of it. Red edges are optimal, but missing in the respective candidate set.}
    \label{fig:d2103-alpha-candidates}
    \vspace*{-6pt}
\end{figure}

Although the subgradient algorithm may mitigate some of these issues, as long as the minimum 1-tree is not the solution to the TSP, it will contain tree-like structures leading to missing edges, as illustrated in \Cref{fig:slkhfail}.
An example of a larger instance where numerous optimal edges are absent in the $\alpha$-candidate set is \texttt{d2103}, which poses a significant challenge for LKH. 
\Cref{fig:d2103-alpha-candidates} highlights the distinctions between an $\alpha$- and a 2-opt-based candidate set. 
Notably, the latter candidate set exhibits increased diversity, especially in long-distance edges.

\begin{figure}[!t]
    \centering
    \includegraphics[width=0.95\linewidth]{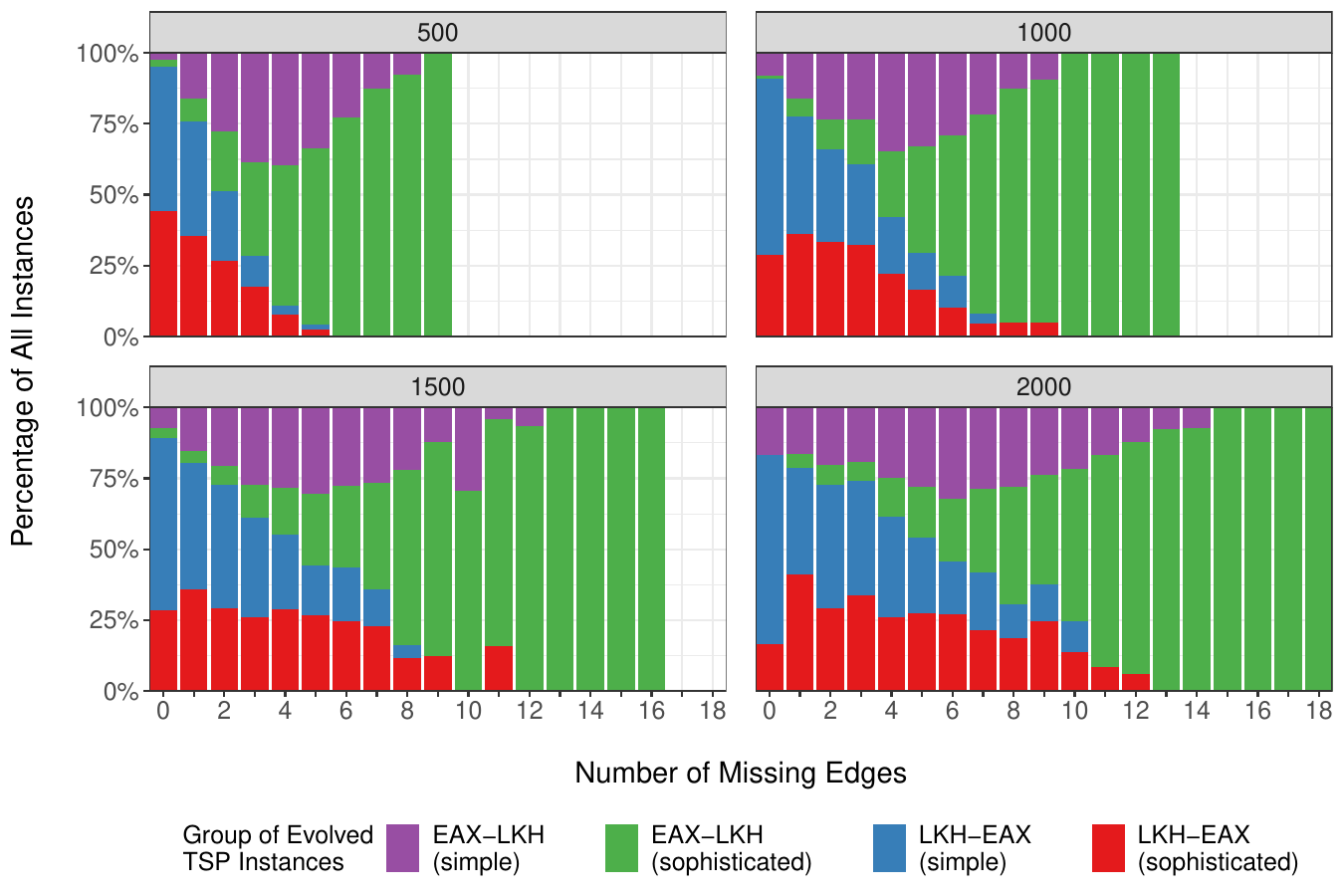}
    \vspace*{-6pt}
    \caption{Share of problem instances with $0$ to $18$ missing edges in the $\alpha$-candidate list (x-axis) for the different groups and instance sizes.}
    \label{fig:alpha-missing-edges}
    \vspace*{-6pt}
\end{figure}

If missing edges can cause LKH to timeout even on a small five-city instance, it is likely that the frequency of optimal edges in the candidate sets for instances intentionally made challenging for LKH correlates with the algorithm's performance.
Previous studies successfully evolved numerous TSP instances that are difficult to solve for LKH but easy for EAX, and vice versa \cite{bossek2019}. However, the authors lacked an explanation for the poor performance. 
To investigate our hypothesis, we illustrate the proportions of instances belonging to one of the evolved instance groups (grouped by the number of missing edges along the x-axis) in \Cref{fig:alpha-missing-edges}. The evolved instance groups are distinguished into four categories, depending on whether they are easy for EAX and difficult for LKH (\texttt{eax-lkh-*}) or vice versa (\texttt{lkh-eax-*}), and whether they were generated using simple or sophisticated mutation operators. 
It is evident that instances without missing edges in the candidate set are predominantly the ones classified as LKH-friendly (i.e., \texttt{lkh-eax-*}).
Additionally, if the amount of missing edges increases, the more likely it is that an instance belongs to the groups of instances that are challenging for LKH (i.e., \texttt{eax-lkh-*}). 
Notably, in cases with many missing edges, the sophisticated evolution strategy dominates, indicating the higher flexibility of this approach. 
This trend remains consistent across all instance sizes, with a roughly linearly scaled increase in the maximal number of missing edges for this small set of sizes. 
In consequence, these findings highlight a significant factor contributing to LKH's poor performance on EAX-friendly instances.

The evolution process identified many scenarios particularly challenging for the minimum 1-tree based pruning strategy, rather than for the inner LKH search procedure. 
This suggests an unfair comparison of LKH in previous algorithm selection studies, based solely on the performance of the $\alpha$-candidate set. 
Restarting EAX, as proposed by \cite{duboislacoste2015}, involves reinitializing the evolutionary algorithm with a new population optimized by 2-opt that may introduce many new edges. 
On the other hand, the construction of the $\alpha$-candidate set is independent of the random seed given to the algorithm, meaning restarting the algorithm will not address the issue of missing edges. 
Therefore, we explore two new restarting strategies: one based on candidates derived from a 2-opt-based population and the other based on a candidate set produced by POPMUSIC.

\section{Experimental Setup}
\label{sec:setup}

The experiments with new LKH configurations were conducted using Intel Xeon Platinum 8470 CPUs, each being repeated for ten folds using a different starting seed ($1\,000\,000 \times$ fold). 
For each restart, the seed was incremented by one. 
The cutoff time is set to one hour per run. 
In accordance with previous studies, we aggregate the run results per instance using the penalized average runtime (PAR10). In case of successful runs, this metric simply computes the average runtime needed to solve the instance. If an instance cannot be solved within the given budget, PAR10 penalizes the corresponding runtime by ten times the maximum runtime, i.e., $36\,000s$, prior to aggregating the runtimes.

We maintain the standard settings of LKH except for the seed and the candidate set. 
In our experiments, we create four distinct restart versions of LKH differentiated by the candidate set construction: LKH$_{\alpha}$, LKH$_{\text{2-opt}}$, LKH$_{\text{pop, fixed}}$, and LKH$_{\text{pop, restart}}$.
LKH$_{\alpha}$ corresponds to the vanilla restart version as described in \cite{duboislacoste2015}, utilizing the $\alpha$-candidate set. 
For LKH$_{\text{2-opt}}$, we provide the 2-opt-based candidate set through a candidate set file. 
This set is generated from $1\,000$ 2-opt-optimized tours, from which all unique edges are symmetrically added to the candidate set and then ordered by their frequency. 
This process results in approximately $5 \times n$ unique edges for all instances, where $n$ represents the instance size. 
Both LKH$_{\text{pop, fixed}}$ and LKH$_{\text{pop, restart}}$ are POPMUSIC-based variants. 
In both variants, we restart the algorithm every time a solution was not found within one run. 
However, for LKH$_{\text{pop, fixed}}$, we reuse a prerecorded POPMUSIC candidate set, whereas for LKH$_{\text{pop, restart}}$, the candidate set is reinitialized at every restart.
Note that when referring to LKH$_{\alpha}$ and EAX, we reuse old performance data \cite{heins2021,heins2023} recorded on other CPUs, which may portray them slightly less favorably. 
However, our focus is on the structural performance differences, specifically whether the algorithm times out on a given instance or not, hence small differences in runtime are negligible. 
Further, note that we do not account for the initialization costs for the same reason. 

All tests were conducted on the $10\,000$ evolved instances from \cite{bossek2019}. 
These instances were chosen to reveal performance variations across different candidate set configurations, as they encompass both highly challenging and easily solvable instances for LKH when using an $\alpha$-candidate set.

\section{Experimental Results}
\label{sec:results}

\begin{figure}[tb]
    \centering
    \includegraphics[width=\columnwidth, trim = 15pt 7pt 20pt 5pt, clip]{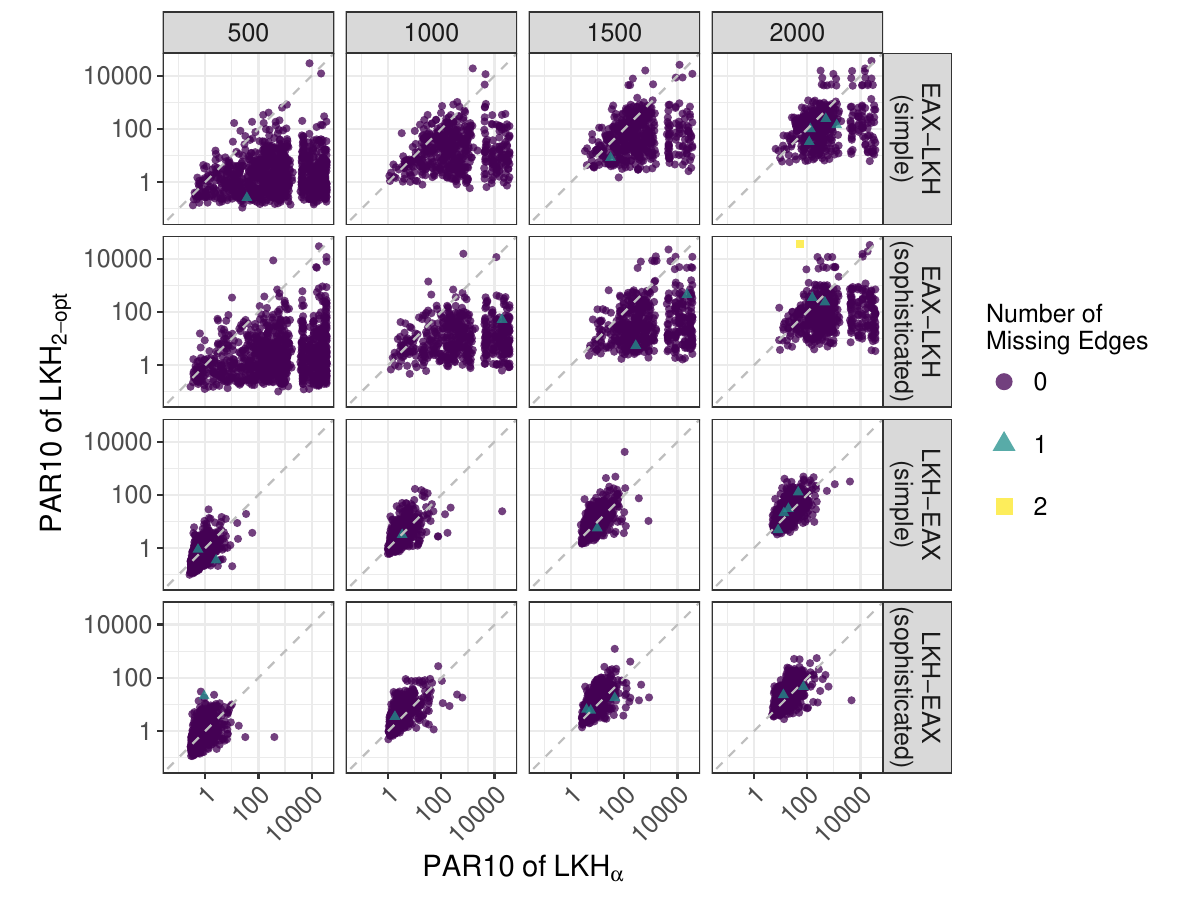}
    \vspace*{-12pt}
    \caption{Log-scaled average PAR10 score of LKH$_{\alpha}$ and LKH$_{\text{2-opt}}$ for each group and size of all evolved problem instances. The color corresponds to how many edges are missing in the 2opt-based candidate set.}
    \label{fig:alpha-vs-twoopt}
    \vspace*{-6pt}
\end{figure}

The candidate sets constructed with the edge frequency information of the 2-opt-optimized Hamiltonian circuits reliably contains all optimal edges. 
Only $26$ of $10\,000$ candidate sets have one missing edge, and only one has two missing edges. 
This leads to a significant performance improvement, with one order of magnitude fewer timeouts observed. 
\Cref{fig:alpha-vs-twoopt} shows detailed performance data of both algorithms, where every point corresponds to one TSP instance.
Instances along the dashed line are solved approximately equally fast by both algorithms. 
Instances below this line are solved faster by LKH$_{\text{2-opt}}$, while instances above are faster solved with LKH$_{\alpha}$. 
The most notable improvement of LKH$_{\text{2-opt}}$ over LKH$_{\alpha}$ is seen in small-sized EAX-friendly instances, where the 2-opt-based variant outperforms the original restart-version on nearly every instance.
Interestingly, the performance on instances specifically designed to be easily solvable by LKH does not show bias towards one of the variants being faster, despite the changed candidate set.
Furthermore, the data supports the observation that a single missing edge in the candidate set does not significantly impact LKH's performance, as the missing edge can be introduced when closing a sequential move.
However, the instance with two missing edges (see red square in the scatterplot of EAX-friendly instances with $2\,000$ cities), is notably more challenging.

\begin{figure}[tb]
    \centering
    \includegraphics[width=\columnwidth, trim = 15pt 7pt 20pt 5pt, clip]{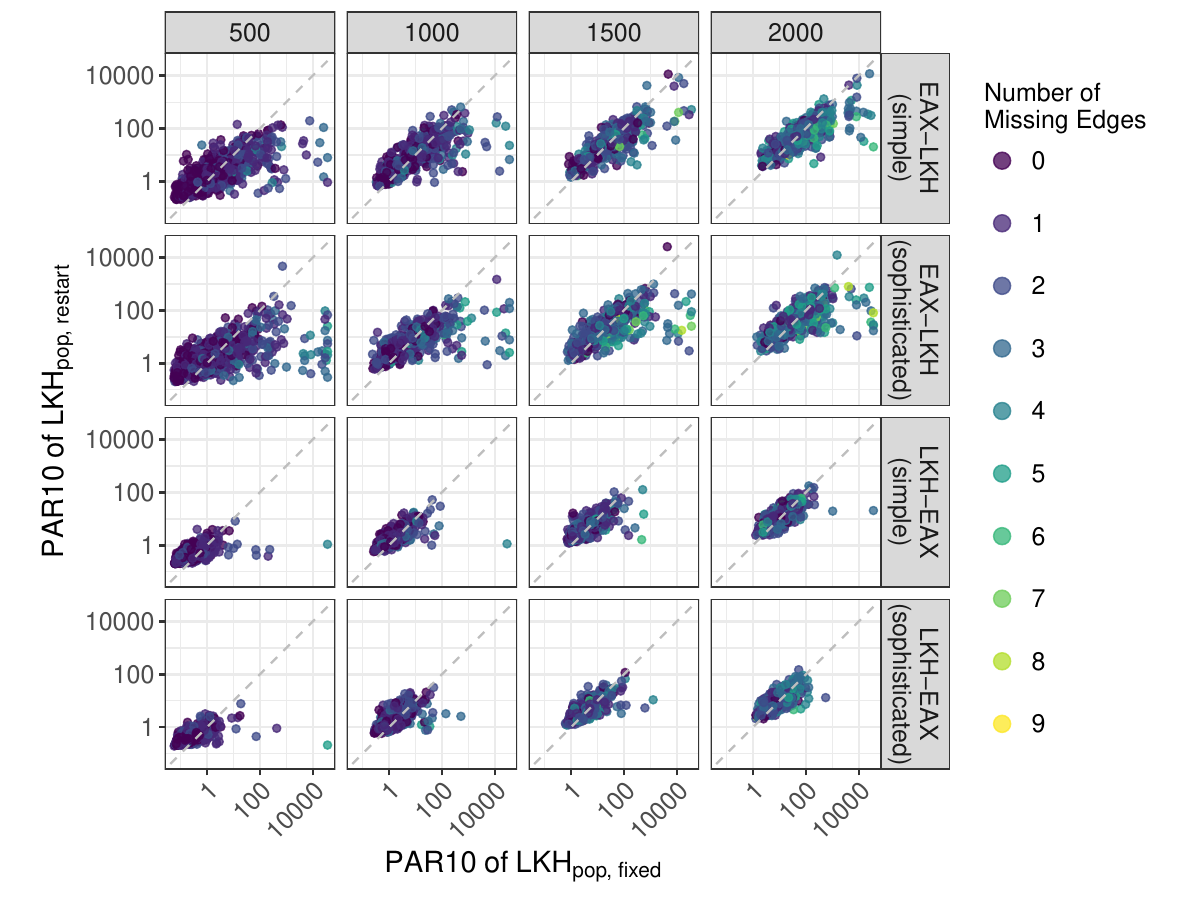}
    \vspace*{-12pt}
    \caption{Log-scaled average PAR10 score of LKH$_{\text{pop, restart}}$ and LKH$_{\text{pop, fixed}}$ for each group and size of all evolved problem instances. The color corresponds to how many edges are missing in the candidate set of LKH$_{\text{pop, fixed}}$.
    }
    \label{fig:popfix-vs-poprest}
    \vspace*{-6pt}
\end{figure}

Although most candidate sets include all required edges and the search restarts with a new initial tour after an unsuccessful run, the candidate set itself may introduce a search bias due to edge ordering. 
To investigate this, we examine the two other new restart-variants based on POPMUSIC: LKH$_{\text{pop, fixed}}$ and LKH$_{\text{pop, restart}}$. 
The performance comparison is shown in \Cref{fig:popfix-vs-poprest}. 
The data clearly indicates that the POPMUSIC-based candidate set misses more important edges (here: up to nine edges per instance). 
Comparing the results across \Cref{fig:alpha-vs-twoopt} and \Cref{fig:popfix-vs-poprest}, it becomes apparent that the LKH-variant based on the fixed POPMUSIC candidate set times out more often compared to LKH$_{\text{2-opt}}$. 
However, including the candidate set in the restart process resolves this issue, and all instances with many missing edges are solved within the time limit. 
Note that most instances leading to timeouts have two or more missing edges. 
However, there are a few instances with at most one missing edges that produced timeouts but could be solved by LKH$_{\text{pop, restart}}$. 
This suggests that resetting the candidate set for every restart can mitigate biases beyond missing optimal edges.

\begin{figure}[tb]
\centering

\includegraphics[width=\linewidth]{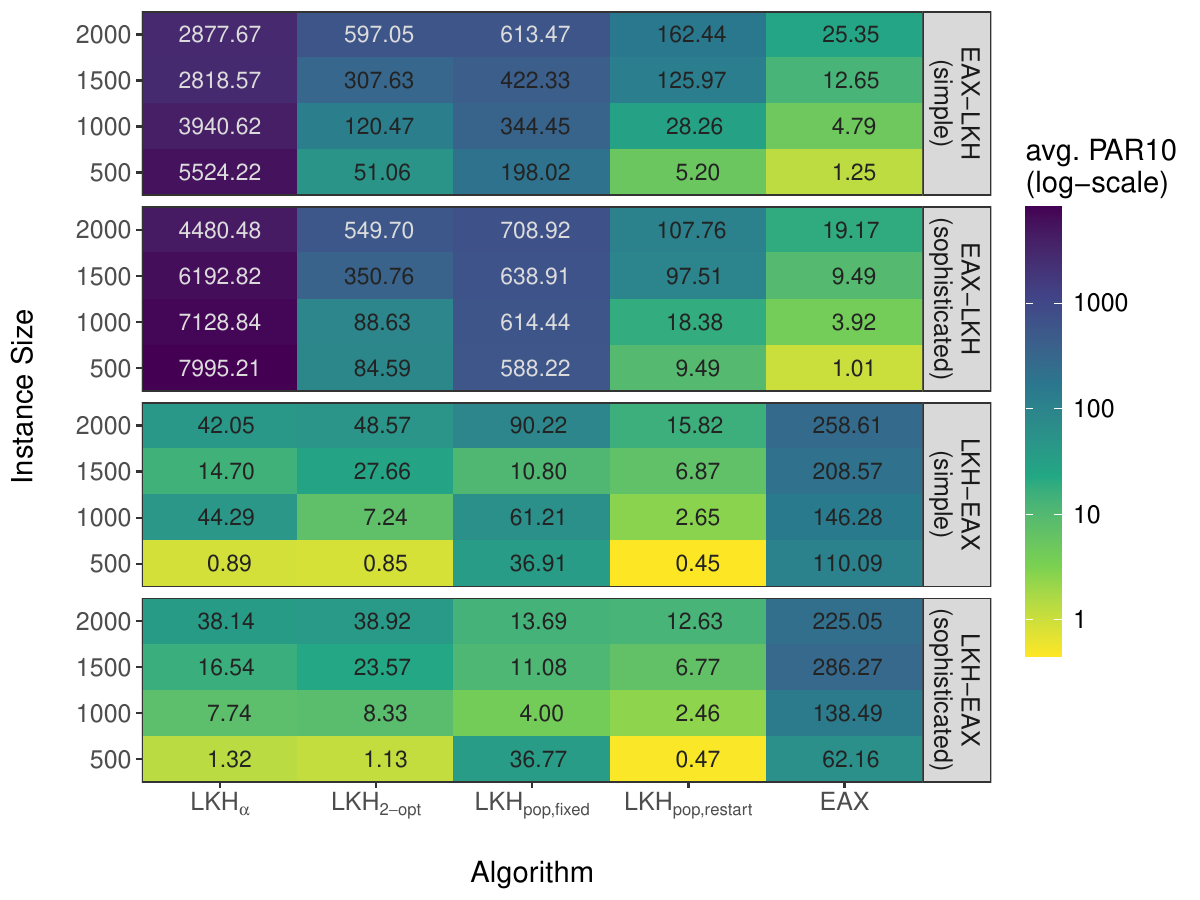}
\caption{Average PAR10-scores of LKH$_{\alpha}$, LKH$_{\text{2-opt}}$, LKH$_{\text{pop,fixed}}$, LKH$_{\text{pop,restart}}$ and EAX for each of the four considered TSP sets and instance sizes. On the EAX-friendly instances (upper two panels), LKH$_{\text{2-opt}}$ strongly outperforms LKH$_\alpha$, while maintaining consistent performance on LKH-friendly instances (lower two panels). Further, LKH$_\text{pop,restart}$ significantly outperforms LKH$_\text{pop,fixed}$ and LKH$_\text{2-opt}$ on all problem categories, highlighting the importance of changing the candidate sets between runs.
}
\label{tab:sum-perf}
\end{figure}

\Cref{tab:sum-perf} provides a summary of the PAR10-performance of all examined LKH-variants across different instance sizes and groups. 
Although the POPMUSIC candidate set was initially designed for large instances, combining it with a restart mechanism yields a much more competitive algorithm compared to the original restart-variant of LKH, which was based on the $\alpha$-candidate set.

\section{Conclusion}
\label{sec:conclusion}

In this study, we explored the impact of various options for the candidate edge set of LKH. 
Our findings reveal that even a small number of missing edges can frequently result in algorithm timeouts. 
Instances that were evolved to be challenging for LKH$_{\alpha}$ were found to be primarily difficult due to the absence of optimal edges in the $\alpha$-candidate set and a lack of changes to the candidate set between restarts. 
This highlights a significant challenge in algorithm selection for the TSP, a topic not thoroughly addressed in prior works. 
Choosing the appropriate algorithm for a specific instance necessitates knowledge of whether optimal edges are absent in the candidate set, which, in turn, requires knowledge of the optimal tour. Yet, such information is obviously a priori not available.

\begin{figure}[!t]
    \centering
    \includegraphics[width=\columnwidth]{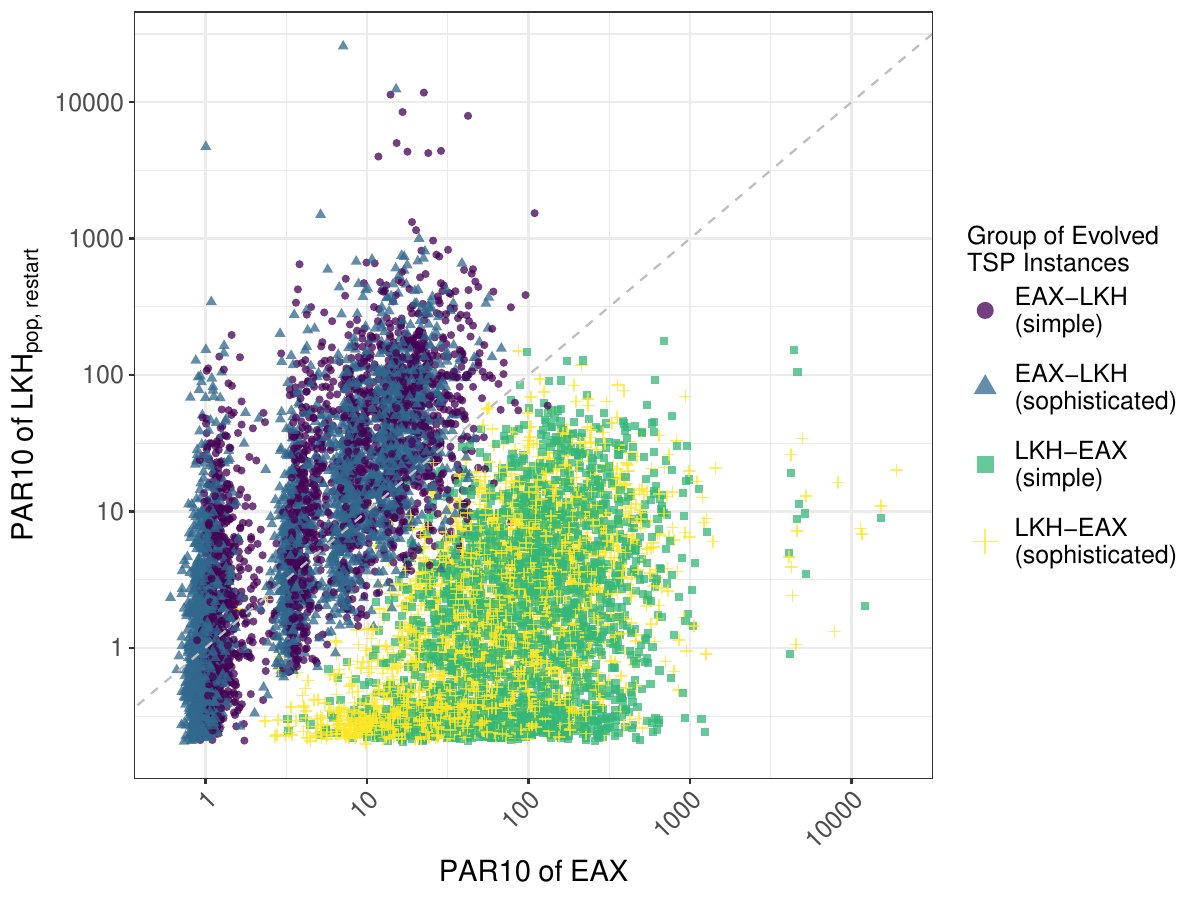}
    \vspace*{-18pt}
    \caption{Log-scaled average PAR10 scores of EAX and LKH$_{\text{pop, restart}}$.}
    \label{fig:eax-vs-poprest}
    \vspace*{-6pt}
\end{figure}

A candidate set generated from 2-opt tours typically includes all optimal edges in the vast majority of cases, leading to notably improved runtimes. 
However, the most optimal performance is achieved with a lightweight restartable candidate set based on Hamiltonian circuits, such as POPMUSIC. 
Our new restart variant, LKH$_{\text{pop, restart}}$, outperforms all other LKH variants and, as shown in \Cref{fig:eax-vs-poprest}, is a true contender to EAX as state-of-the-art TSP solver. 

These findings open up numerous avenues for further research. 
Algorithm selectors can now benefit from true performance complementarity between EAX and our proposed solver, and thereby leverage the substantial performance improvement without facing the challenge of (avoiding) excessive timeouts for LKH. 
Also, the candidate set creation with POPMUSIC can be further optimized, and the resulting performance differences can be examined depending on the underlying instance types. 
New instances can be evolved to gain further insight into the complementary nature of EAX and LKH. 
Lastly, the impact of initial tours on EAX warrants investigation as well, considering the relatively long initialization time of the 2-opt population and the potential analogous enhancements with a different initial population type.

\begin{credits}

\subsubsection{\ackname}

The authors gratefully acknowledge the computing time made available to them on the high-performance computer at the NHR Center of TU Dresden. This center is jointly supported by the Federal Ministry of Education and Research and the state governments participating in the NHR (\url{www.nhr-verein.de/unsere-partner}).
This research received financial support from the German Academic Exchange Service (DAAD) under the Program for Project-Related Personal Exchange (PPP).
Lennart Schäpermeier and Pascal Kerschke acknowledge support by the \href{https://scads.ai}{\em Center for Scalable Data Analytics and Artificial Intelligence (ScaDS.AI) Dresden/Leipzig}.

\end{credits}
%
%
%
\bibliographystyle{splncs04}
\bibliography{library}

\end{document}